\documentclass[letterpaper, 12pt]{article}
\usepackage[left=1in, right=1in, top=1in, bottom=1in]{geometry}
\usepackage{times}
\usepackage{graphicx}
\usepackage{amsmath, amssymb}
\usepackage{cite}
\usepackage{caption}
\usepackage{titlesec}
\usepackage{booktabs}
\usepackage{array}
\usepackage{float}
\usepackage{titlesec}

\titleformat{\section}{\normalfont\Large\bfseries}{\thesection.}{1em}{}
\titleformat{\subsection}{\normalfont\large\itshape}{\thesubsection.}{1em}{}
\titleformat{\subsubsection}{\normalfont\normalsize\bfseries}{\thesubsubsection.}{1em}{}

\title{Evaluating the Suitability of Different Intraoral Scan Resolutions for Deep Learning-Based Tooth Segmentation}

\author{
        Daron Weekley$^{1}$, Jace Duckworth$^{1}$, Anastasiia Sukhanova$^{1}$, Ananya Jana$^{1}$,\\
        $^{1}$Department of Computer Sciences and Electrical Engineering, \\ Marshall University, Huntington, WV, USA \\
        Corresponding author: {\tt jana@marshall.edu}
}

\date{}
\begin{document}
\raggedright
\maketitle
\thispagestyle{empty}
\pagestyle{empty}

\section*{Abstract}
%\begin{abstract}
Intraoral scans are widely used in digital dentistry for tasks such as dental restoration, treatment planning, and orthodontic procedures. These scans contain detailed topological information, but manual annotation of these scans remains a time-consuming task. Deep learning-based methods have been developed to automate tasks such as tooth segmentation. A typical intraoral scan contains over 200,000 mesh cells, making direct processing computationally expensive. Models are often trained on downsampled versions, typically with 10,000 or 16,000 cells. Previous studies suggest that downsampling may degrade segmentation accuracy, but the extent of this degradation remains unclear.  Understanding the extent of degradation is crucial for deploying ML models on edge devices. This study evaluates the extent of performance degradation with decreasing resolution.  We train a deep learning model (PointMLP) on intraoral scans decimated to 16K, 10K, 8K, 6K, 4K, and 2K mesh cells. Models trained at lower resolutions are tested on high-resolution scans to assess performance. Our goal is to identify a resolution that balances computational efficiency and segmentation accuracy. 

%\end{abstract}

% \begin{keywords}
% Keywords: tooth segmentation, intraoral scan, deep learning, mesh processing, point cloud, dental mesh, mesh resolution, multiresolution mesh, decimated mesh
% \end{keywords}

%
\section*{Introduction}
Intraoral scanning is a digital technique that captures a 3D representation of an intraoral region. These scans are widely used in digital dentistry for dental restoration, orthodontic treatment planning, smile design, and surgery. However, many of these tasks still require manual processing, such as annotation. Annotating a single scan can take 45 to 60 minutes. Hence, automatic methods are necessary to reduce the manual efforts.

There are many deep learning-based methods that have been developed for the task of tooth segmentation, dental crown generation, and dental alignment, often adapting techniques from general point cloud processing. Intraoral scans can contain over 200,000 mesh triangles, making direct processing computationally prohibitive. The existing works downsample the original scans to a lower resolution such as 16K or 10K mesh cells as shown in Table.~\ref{tab:downsampled_data}. Prior studies \cite{zanjani2019deep, chen2024deep} suggest that downsampling degrades the fine tooth features (for example, curvature information), but a systematic evaluation of its impact on segmentation accuracy has not been done. Our study attempts to fill in this gap by analyzing the effect of different resolution levels on model performance.

 Deep-learning models trained on downsampled meshes are more suitable for deployment on devices that require methods with lower memory footprint. However, downsampling beyond a certain threshold can reduce segmentation accuracy. 
In this study, we evaluate tooth segmentation performance when downsampling intraoral scans across different resolutions. Using PointMLP \cite{ma2022rethinking}, a deep learning model for point cloud processing, we assess segmentation accuracy on 10K and 16K scans. 
By evaluating segmentation performance across various resolutions, this study provides insights into the feasibility of using downsampled intraoral scans for deep learning-based segmentation.
\begin{table}[h]
\centering
\begin{tabular}{|c|c|}
\hline
\textbf{Author/Paper} & \textbf{Faces/Points} \\
\hline
Xu et al. \cite{xu2018} & 40,000 points \\
\hline
Sun et al. \cite{sun2020} & 10,000 points \\
\hline
Zhang et al. \cite{zhang2019} & 16,000 points \\
\hline
Cui et al. \cite{cui2019} & 10,000 cells \\
\hline
Zhang et al. \cite{zhang2020} & 16,000 cells \\
\hline
Tian et al. \cite{tian2020} & 16,000 cells \\
\hline
Zanjani et al. \cite{zanjani2020} & 16,000 points \\
\hline
Wu et al. \cite{wu2023} & 10,000 cells \\
\hline
Lian et al. \cite{lian2021} & 10,000 points \\
\hline
Zhang et al. \cite{zhang2023} & 10,240 points \\
\hline
Benhamadou et al. \cite{benhamadou2023} & 10,000 points \\
\hline
Jana et al. \cite{jana2023} & 16,000 points \\
\hline
\end{tabular}
\caption{Intraoral scan resolutions used in various studies}
\label{tab:downsampled_data}
\end{table}

Current deep learning-based approaches have shifted toward data-driven feature extraction. Xu et al.\cite{xu20183d} introduced a multi-stage framework using CNNs for mesh cell labeling. Tian et al.\cite{tian2019automatic} employed octree partitioning followed by a 3D CNN for hierarchical tooth segmentation. Jana et al. \cite{jana20233d} developed a dual-branch segmentation method, leveraging sparse surface representation and curvature learning. 3D tooth segmentation methods primarily rely on fully supervised learning, though some, such as DArch \cite{qiu2022darch} and TSegNet \cite{cui2021tsegnet}, use weak or semi-supervised learning. As such, these approaches typically operate on 16K or 10K mesh cells per scan. To our knowledge, no study systematically evaluates the impact of downsampling on tooth segmentation performance.

\section{Methods}

\subsection{PointMLP}
\begin{figure}[H]
\centering
\includegraphics[width=1.0\textwidth]{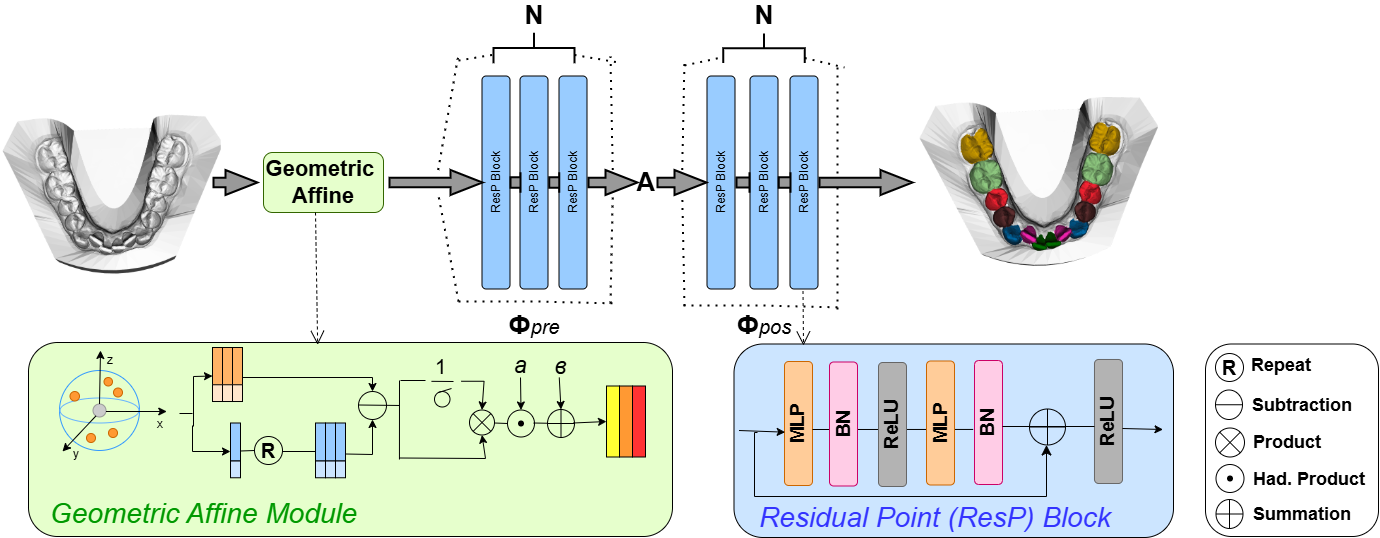}
\caption{PointMLP method comprises of residual blocks ($\phi_{pre}$ and $\phi_{post}$ ). $\alpha$ and $\beta$ are the learnable parameters. $\sigma$ denotes the feature deviation across all groups. The PointMLP method transforms the points via a normalization operation while maintaining original geometric
properties. (Zoom in for better visibility)}
\label{fig:diff_res_diagram}
\end{figure}

In this work, we have used a point cloud segmentation method named PointMLP\cite{ma2022rethinking}. The PointMLP method uses a geometric affine module, which has demonstrated exceptional capability of capturing local geometry, even from sparse representation. Due to these characteristics, the PointMLP method seems to be a reasonably good choice. Given an input point cloud, PointMLP extracts local features step by step using residual point MLP blocks. First, a geometric affine module transforms the local points. Features are then extracted both before and after the aggregation process. By repeating these stages, PointMLP expands the receptive field and captures the full geometric structure of the point cloud. Figure ~\ref{fig:diff_res_diagram} shows a visualization of the PointMLP method.

\subsection{Experimental Set-up} 
\begin{figure}[H]
\centering
\includegraphics[width=0.4\textwidth]{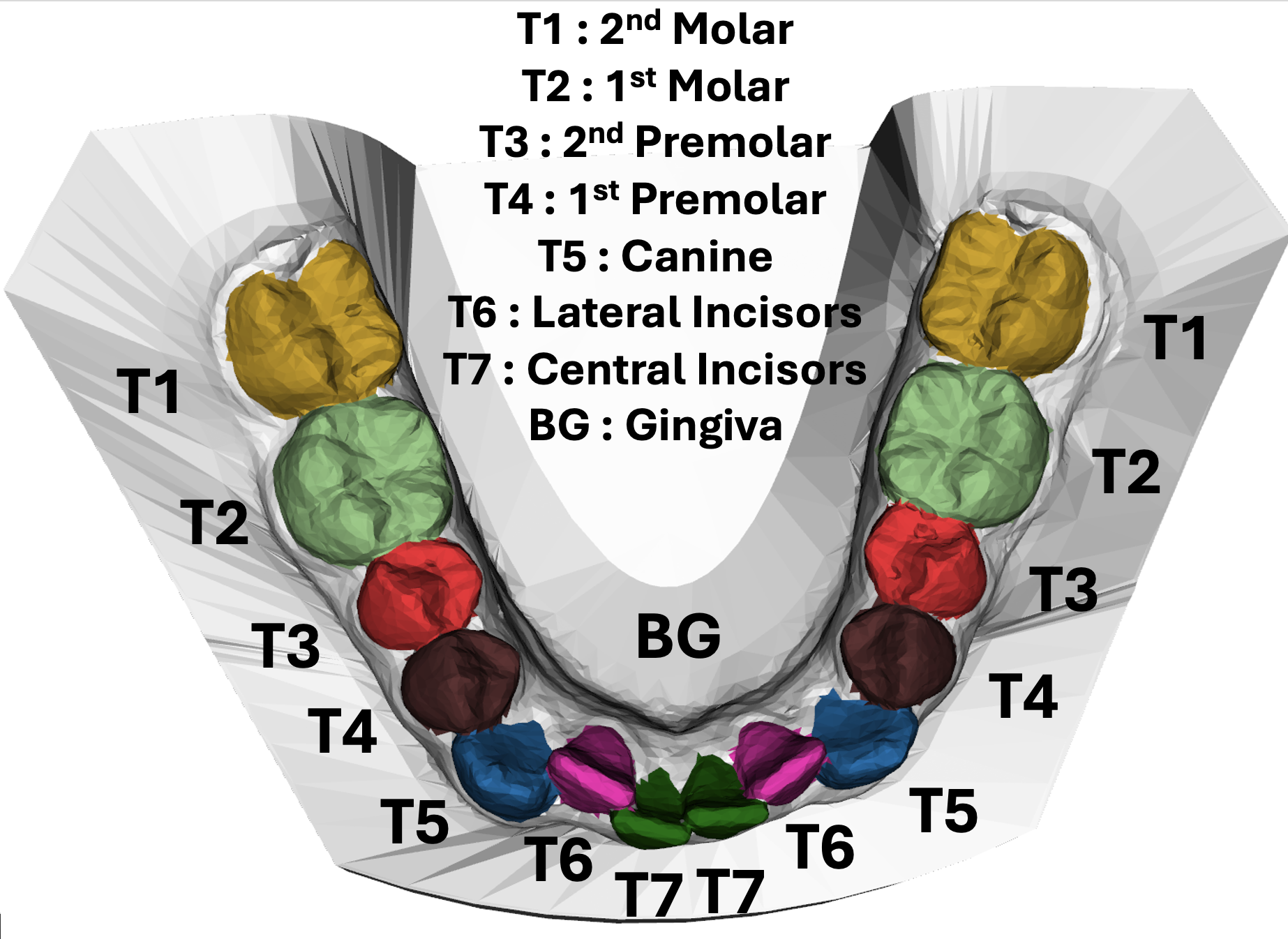}
\includegraphics[width=0.58\textwidth]{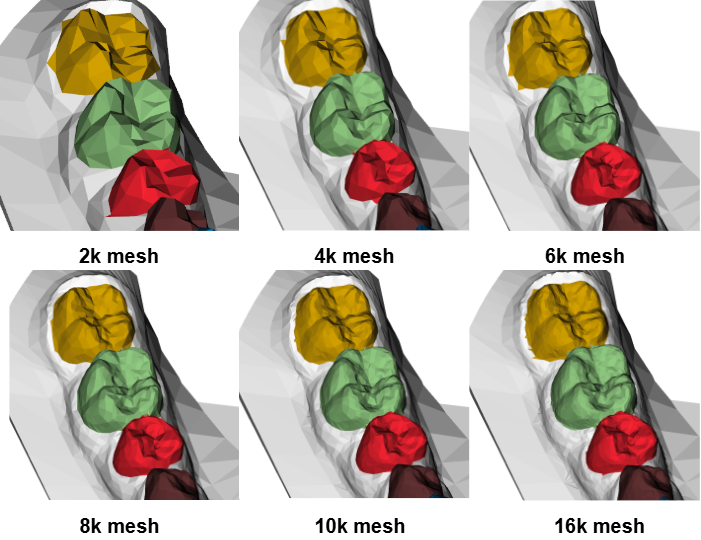}
\caption{A labeled tooth mesh (left) and different resolutions of the same region in the intraoral scan.(Zoom in for better visibility)}
\label{fig:diff_res}
\end{figure}

The original intraoral scans ($\sim$200K cells) were downsampled to the resolutions of 2K, 4K, 6K, 8K, 10K and 16K and a separate model was trained for each resolution. The results of the prediction were upsampled via KNN method to the 10K and 16K resolution to understand the effectiveness of the models trained at lower resolutions (2K, 4K, 6K and 8K). Examples of the differences in resolutions are shown in Figure~\ref{fig:diff_res}.

\subsection{Dataset \& Evaluation Metrics}
We use the public dataset \textit{3D Teeth Seg Challenge 2022}\cite{ben2022teeth3ds}. The intraoral scans provided the ground truth annotations of different teeth and gum regions, including central incisors (T7), lateral incisors (T6), canine/cuspids (T5), 1st premolars (T4), 2nd premolars (T3), 1st molars (T2), 2nd molars (T1), and background/gingiva (BG)(Figure~\ref{fig:diff_res}). Tooth segmentation is the task of separating 3D dental models into these unique semantic parts.
We use Dice Score (DSC), Overall Accuracy (OA), Sensitivity (SEN) and Positive Predictive Value (PPV) to evaluate the performance of our model.

\subsection{Data Pre-Processing}
In our experiments, we used a subset of the public dataset consisting of 571 subjects from the training set of the dataset. The lower jaw scans of the subjects with a maximum of 14 teeth were selected.
Each of these resolutions have been created from the original mesh by the process of mesh quadric decimation.  
Each mesh cell can be described with four points - three vertices of the mesh triangle and the barycenter of the mesh triangle along with the normals at each of these four points. With these four points, a 24 dimensional vector is constructed comprising 12 coordinate vectors and 12 normal vectors at the four points, respectively. The intraoral scans have a base which is not part of the gum.  In our current work, we crop a portion of the dental mesh from the scan. Afterward, we manually verified the meshes to ensure that we have teeth and gingiva for all the models. 
\subsection{Data Augmentation}
For better generalizability of the model, we augment the training and validations sets by combining 1) random rotation, 2) random translation, and
3) random rescaling of each 3D dental surface in reasonable ranges. Specifically, along each of the three axes in the 3D space, a training/validation surface has 50\% probability of translation with a displacement and zoom with a ratio uniformly sampled between [\textminus10, 10] and [0.8, 1.2], respectively. Each training/validation surface has 50\% probability to be rotated along the x, y, and z\textminus axes with angles uniformly sampled between [\textminus$\pi$, $\pi$]. The combination of these random operations simulated 4 “new” cases from each original surface. We split the data at an 80:20 ratio for training and testing. Next, we allocate 20\% of the training data for the purpose of validation.

%%%%%%%%%%%%%%%%%%%%%%%%%%%%%%%%%%55

\subsection{Training and Evaluation}
We train the segmentation model using PointMLP with multi-resolution intraoral scans. Training is conducted for 200 epochs using the Adam optimizer with an initial learning rate of 0.001. A StepLR scheduler reduces the learning rate by a factor of 0.5 every 120 epochs. The batch size is set to 16 and each epoch took approximately 5 minutes to complete.
Training is conducted on RTX 6000 GPUs. The entire pipeline, including preprocessing, training, and evaluation, is implemented in PyTorch.

\section{Results and Analysis}

\subsection{Segmentation Performance}
\begin{figure}[H]
\centering
\includegraphics[width=1\textwidth]{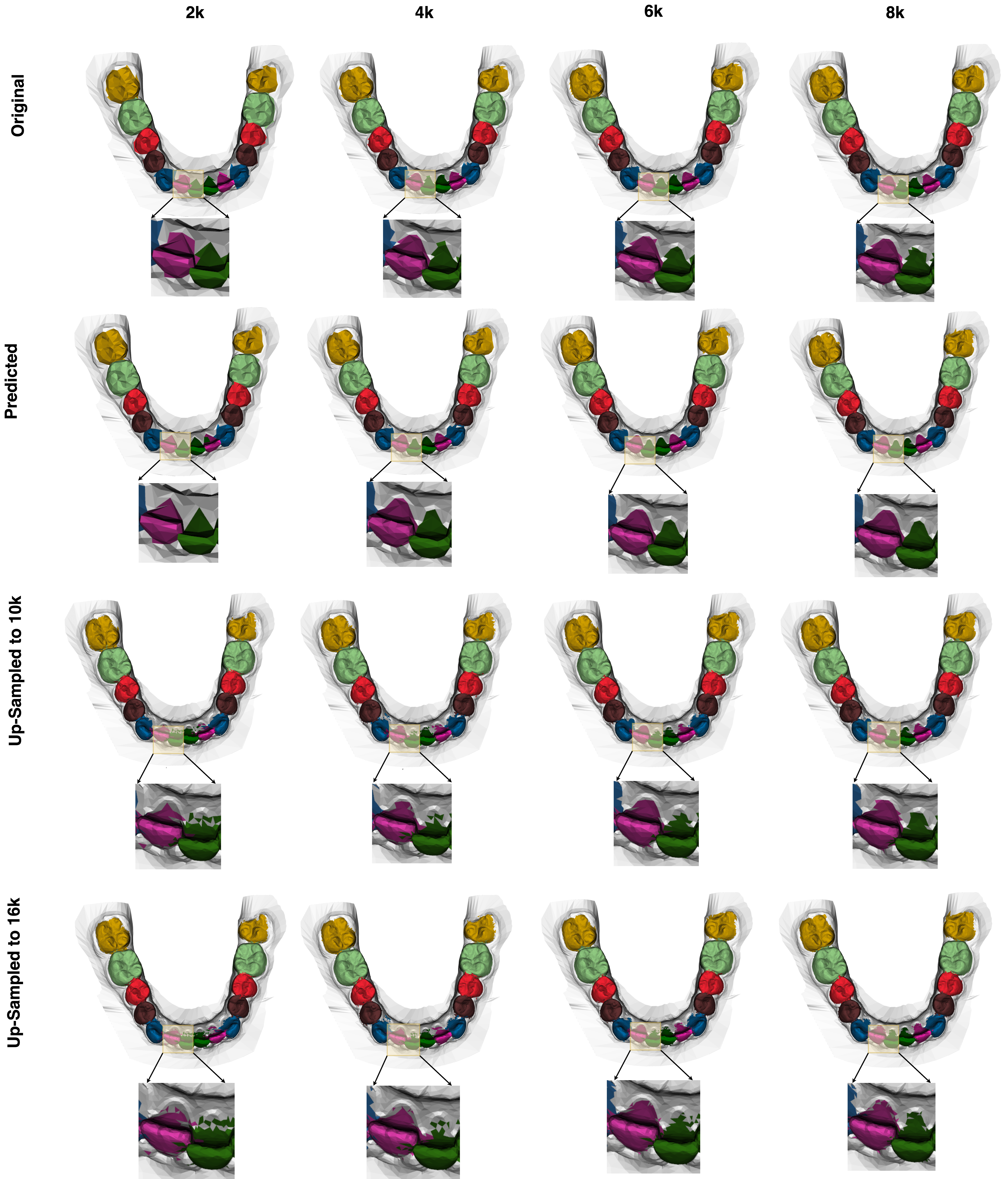}
\caption{Qualitative results of predictions from the original models and their upsampled versions. (Zoom in for better visibility)}
\label{tooth_model_1}
\end{figure}

\begin{figure}[H]
\centering
\includegraphics[width=1\textwidth]{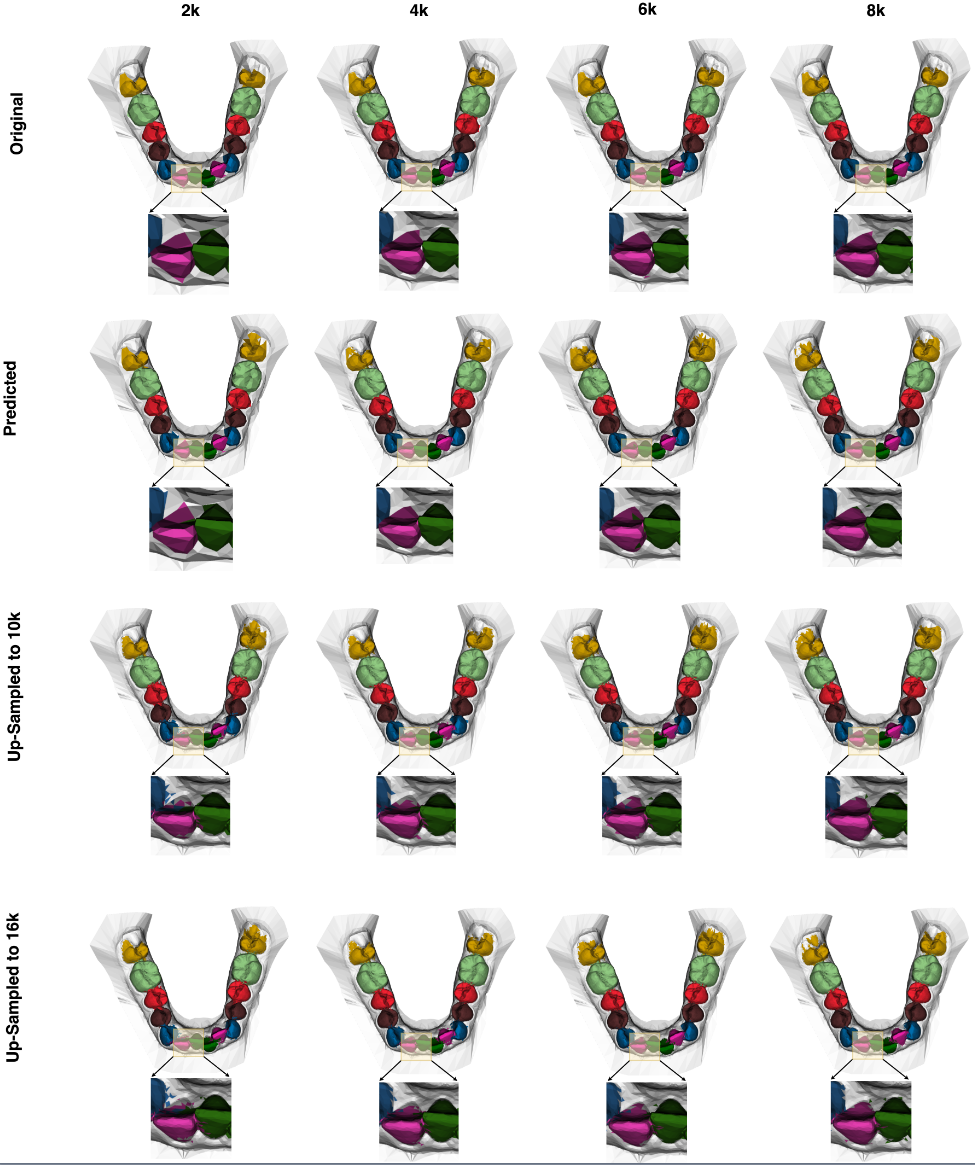}
\caption{Qualitative results of predictions from the original models and their upsampled versions (continued). (Zoom in for better visibility) }
\label{tooth_model_2}
\end{figure}

\begin{figure}[H]
\centering
\includegraphics[width=1\textwidth]{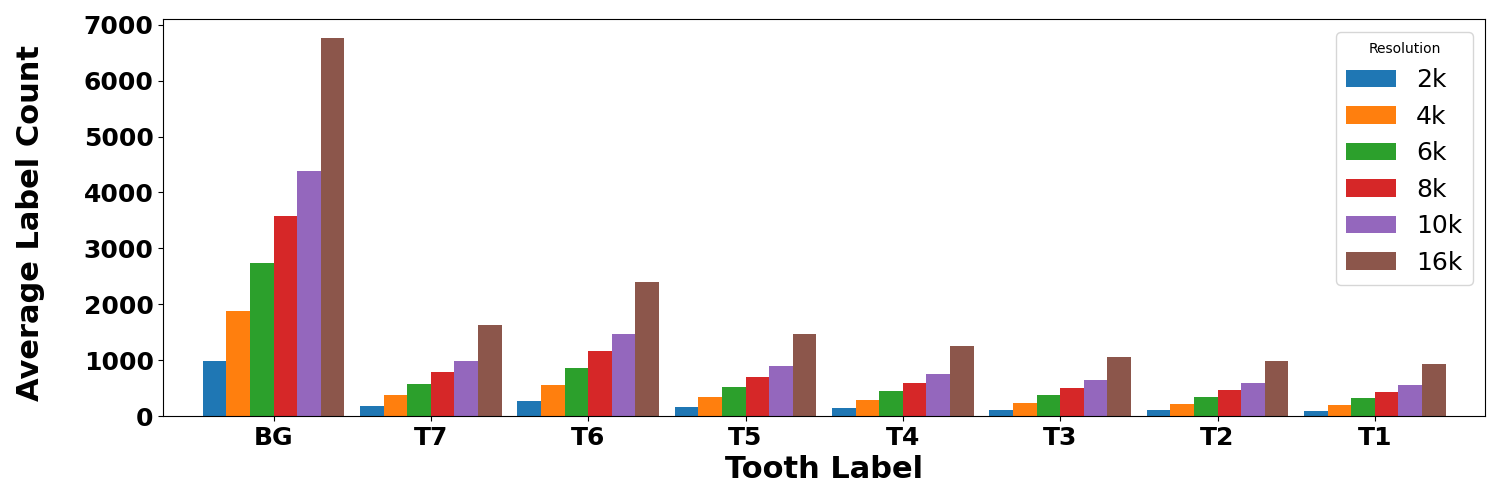}
\caption{The distribution of the labels across different classes. (Zoom in for better visibility)}
\label{tooth_distr_label}
\end{figure}

We evaluate the segmentation performance of PointMLP across different resolutions and report the results in Table~\ref{tab:per_class_acc}. The distribution of the segmentation labels is shown in Figure~\ref{tooth_distr_label} The model trained at each resolution is tested against the standard 10K and 16K resolution scans to analyze performance degradation. The overall results from these experiments are listed in Table~\ref{tab:allres_upsampled}. We analyze segmentation performance for background (BG) and each tooth class (T1–T7) and report the Dice scores in Table~\ref{tab:allres_upsampled}. We measure per-class segmentation accuracy and report the results in Table~\ref{tab:per_class_acc}. The qualitative results of our experiments is shown in Figure~\ref{tooth_model_1} and Figure~\ref{tooth_model_2}. 

\vspace{12 cm}

\begin{table*}[t]
\centering
\caption{Per-class DSC score across different resolutions and their upsampled versions.}
\begin{tabular}{|c|c|c|c|c|c|c|c|c|}
\hline
\textbf{Resolution} & \textbf{BG} & \textbf{T1} & \textbf{T2} & \textbf{T3} & \textbf{T4} & \textbf{T5} & \textbf{T6} & \textbf{T7} \\ \hline
16K  & 0.9646 & 0.9566 & 0.9457 & 0.9443 & 0.9402 & 0.9265 & 0.9163 & 0.8606 \\ \hline
10K  & 0.9701 & 0.9605 & 0.9504 & 0.9500 & 0.9442 & 0.9267 & 0.9171 & 0.8435 \\ \hline
8K   & 0.9728 & 0.9642 & 0.9527 & 0.9527 & 0.9440 & 0.9252 & 0.9232 & 0.8441 \\ \hline
8K (to 16K) & 0.9468 & 0.9181 & 0.9164 & 0.9068 & 0.8964 & 0.8929 & 0.8927 & 0.8389 \\
        \hline
8K (to 10K) & 0.9569 & 0.9283 & 0.9256 & 0.9121 & 0.9070 & 0.9048 & 0.9013 & 0.8422 \\
        \hline
6K   & 0.9752 & 0.9686 & 0.9573 & 0.9564 & 0.9488 & 0.9250 & 0.9177 & 0.8533 \\ \hline
6K (to 16K) & 0.9405 & 0.9112 & 0.9075 & 0.8999 & 0.8873 & 0.8784 & 0.8777 & 0.8344 \\
        \hline
6K (to 10K) & 0.9502 & 0.9209 & 0.9155 & 0.9044 & 0.8965 & 0.8885 & 0.8849 & 0.8373 \\
        \hline
4K   & 0.9786 & 0.8793 & 0.9430 & 0.9437 & 0.9738 & 0.9673 & 0.9566 & 0.9680 \\ \hline
4K (to 10K) & 0.8319 & 0.7536 & 0.6679 & 0.6626 & 0.6498 & 0.6426 & 0.6145 & 0.5569 \\
        \hline
4K (to 16K) & 0.8208 & 0.7510 & 0.6647 & 0.6585 & 0.6481 & 0.6397 & 0.6116 & 0.5558 \\
        \hline
2K   & 0.9738 &	0.8760 &	0.9410 &	0.9347 & 0.9636 & 0.9495	& 0.9371 &	0.9504
 \\ \hline
 2K (to 10K)  & 0.8213 & 0.7578 & 0.6905 & 0.6838 & 0.6546 & 0.6408 & 0.6303 & 0.5651 \\
     \hline
2K (to 16K)  & 0.8320 & 0.5697 & 0.6923 & 0.6435 & 0.6876 & 0.6571 & 0.6317 & 0.7586 \\
        \hline
\end{tabular}
\label{tab:per_class_acc}
\end{table*}

\begin{table}[htbp]
\centering
\caption{Tooth segmentation results using PointMLP with different resolutions}
\begin{tabular}{|c|c|c|c|c|c|}
\hline
Input Size (\#mesh cells) & OA & DSC & SEN & PPV & Inference Time (ms) \\ \hline
16K  & 0.9476 & 0.9318 & 0.9451 & 0.9353 & 158.93\\ \hline
10K  & 0.9515 & 0.9328 & 0.9467 & 0.9377 & 248.78\\ \hline

8K  & 0.9532 & 0.9348 & 0.9495 & 0.9396 & 207.89\\ \hline
8K (upsample to 10K) & 0.2454 & 0.9097 & 0.9172& 0.9210 & 258.26\\ \hline
8K (upsample to 16K) &  0.2454 & 0.9011 &  0.9061& 0.9150 & 209.05\\ \hline
6K & 0.9549 & 0.9377 & 0.9490 & 0.9442 & 163.45 \\ \hline
6K  (upsample to 10K)  & 0.2372 & 0.8997 & 0.9015 & 0.9166 & 298.59 \\ \hline
6K  (upsample to 16K)  & 0.2372 & 0.8921 & 0.8916 & 0.9115 & 190.06 \\ \hline
4K  & 0.9642 & 0.9513 & 0.9552 & 0.9591 & 279.69\\ \hline
4K (upsample to 10K)  & 0.2575 &0.6724 & 0.6836 & 0.7340 & 302.09\\ \hline
4K (upsample to 16K)  & 0.2575 &0.6687 & 0.6788 & 0.7317 & 192.33\\ \hline
2K  & 0.9577 & 0.9408 & 0.9465 & 0.9491 & 102.22\\ \hline
2K (upsample to 10K)  & 0.2956 & 0.6840 & 0.6942 & 0.7400 & 233.36\\ \hline
2K (upsample to 16K)  & 0.2956 & 0.6805 & 0.6904 & 0.7376 & 413.72\\ \hline
\end{tabular}
\label{tab:allres_upsampled}
\end{table}

\begin{figure}[H]
\centering
\includegraphics[width=1\textwidth]{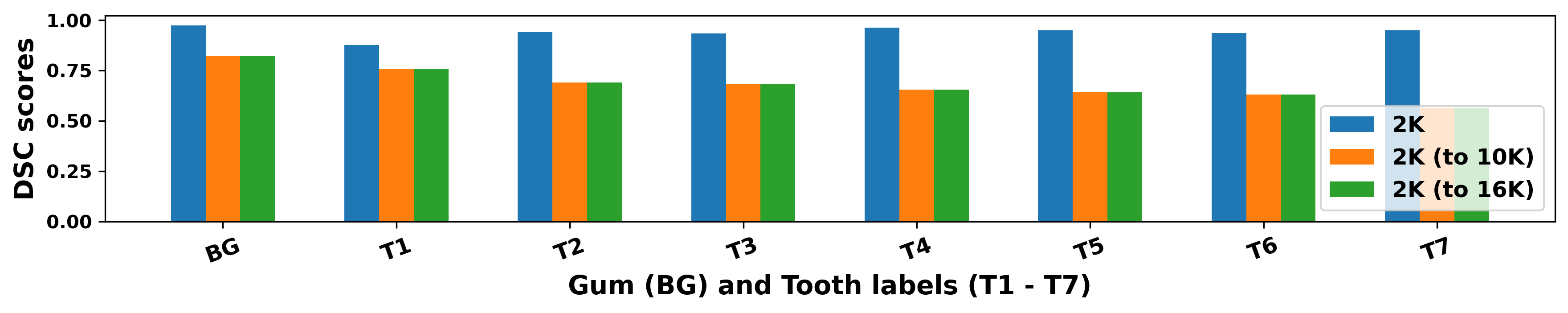}
\includegraphics[width=1\textwidth]{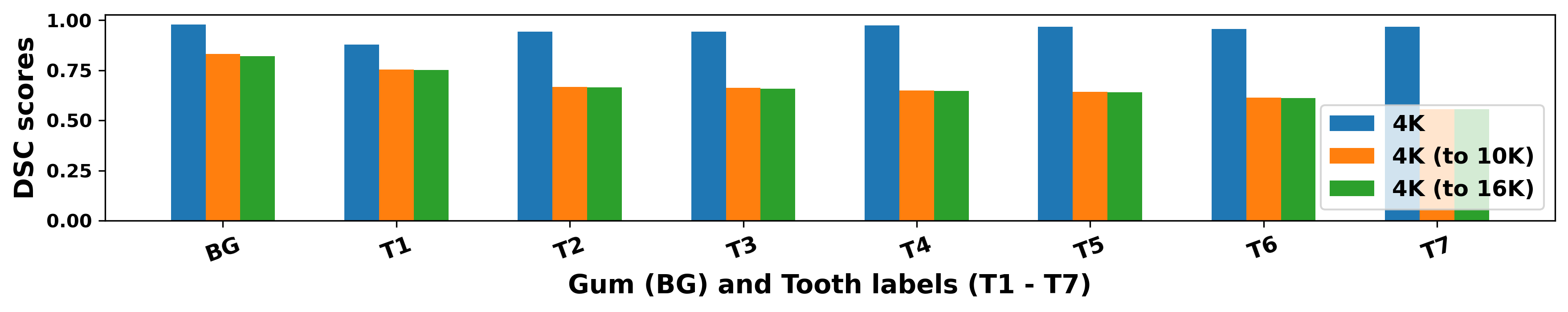}
\includegraphics[width=1\textwidth]{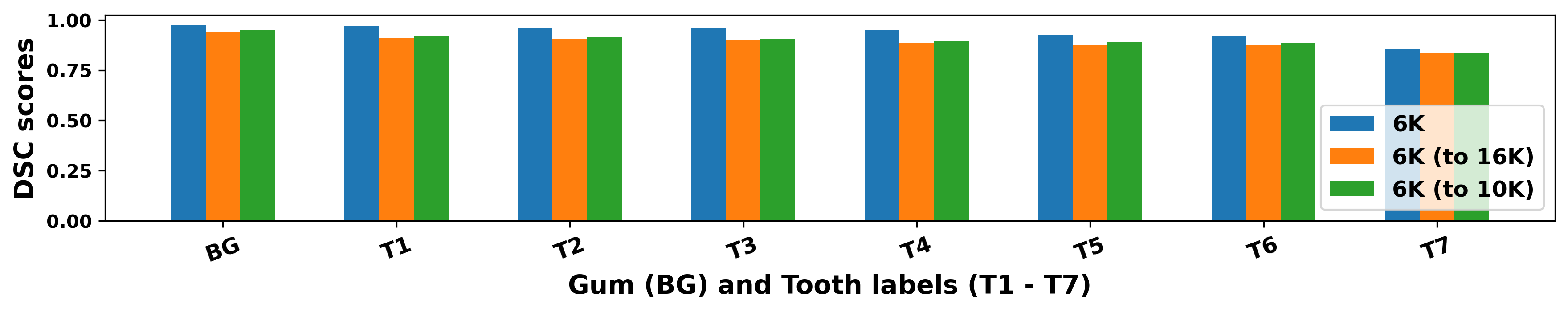}
\includegraphics[width=1\textwidth]{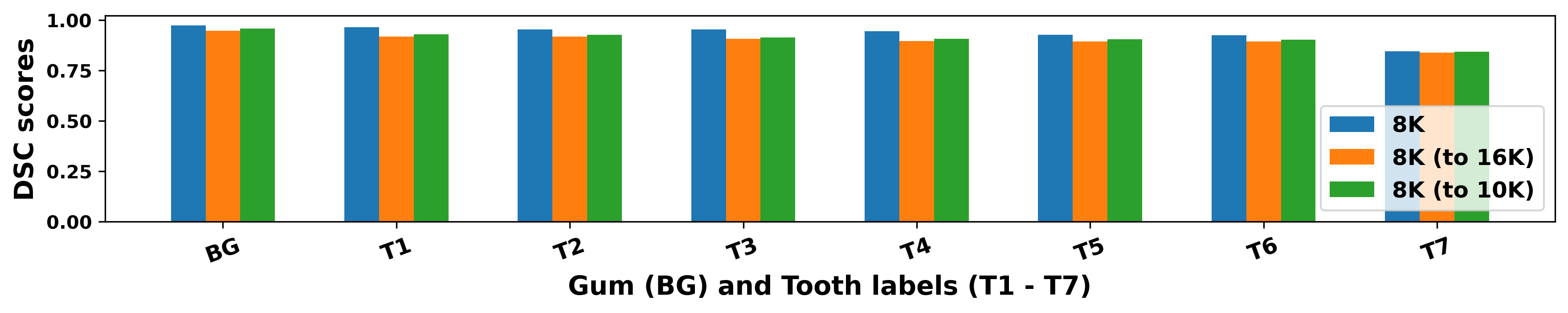}
\caption{A graphical visualization of per-class DSC score across different resolutions compared to their upsampled cohorts. Each resolution and corresponding upsampled version is represented by one of the four graphs and DSC scores are compared.}
\label{tooth_distr}
\end{figure}
\section{Discussion}
There is a clear visual difference between the labels of the decimated intraoral scans and the predicted labels from the model trained on that resolution as can be seen from Figures~\ref{tooth_model_1},~\ref{tooth_model_2}. The predicted labels appear to correct for errors present in the original decimated mesh labels at all resolutions. These corrections result in cleaner models with well-defined labels that are more accurately positioned than their decimated counterparts.
We also observe that the performance of the lower resolution model when upsampled to 10K and 16K show only slight differences in DSC scores as shown in Fig.~\ref{tooth_distr} and Table~\ref{tab:allres_upsampled}. Our experiments showed that the lower resolutions, such as 6K, can be still considered for edge devices as the predictions retain performance $\sim$90 even when upsampled to 10K or 16K models.

\section{Conclusion and Limitations}

This study analyzed how intraoral scan resolution affects the performance of deep learning-based tooth segmentation. The results indicate that while models trained on lower resolutions retain reasonable segmentation accuracy, performance degradation becomes significant at resolutions below 6K. Background gingiva segmentation remain stable across resolutions, but fine-grained tooth boundary recognition deteriorates at lower resolutions. Certain tooth classes, particularly premolars and canines, exhibit higher misclassification rates when trained on lower resolution meshes.

We acknowledge that the experiments were limited to a single segmentation architecture, PointMLP. For better generalization, more experiments need to be conducted. Our study focuses on only a particular aspect, which is the trade-off between resolution and segmentation accuracy. Although, the validity of these findings needs to be investigated in clinical settings, we were able to provide a preliminary quantitative assessment of segmentation performance across different intraoral scan resolutions. 

\bibliographystyle{IEEEtran}
\bibliography{refs}

\end{document}